\documentclass[preprint,10pt,hyperref]{elsarticle}

\usepackage{times}
\usepackage{latexsym}
\usepackage{graphicx}
\usepackage{algorithm}
\usepackage{multirow}
\usepackage{amsmath}
\usepackage{amsthm}
\usepackage{amsfonts}
\usepackage{algorithmic}
\usepackage{booktabs}
\usepackage{caption}
\usepackage{stfloats}
\usepackage{fullpage}

\begin{document}
\begin{frontmatter}

\title{TDRE: A Tensor Decomposition Based Approach for Relation Extraction}
\author[lab1]{Bin-Bin Zhao}
\ead{zbbuestc@sina.com}
\author[lab1]{Liang Li\corref{cor1}}
\ead{plum\_liliang@uestc.edu.cn, plum.liliang@gmail.com}
\author[lab1]{Hui-Dong Zhang}

\address[lab1]{School of Mathematical Sciences, University of
               Electronic Science and Technology of China, 611731,
               Chengdu, P.R. China}

\cortext[cor1]{Corresponding author}

\begin{abstract}
Extracting entity pairs along with relation types from unstructured texts is a fundamental subtask of information extraction. Most existing joint models rely on fine-grained labeling scheme or focus on shared embedding parameters. These methods directly model the joint probability of multi-labeled triplets, which suffer from extracting redundant triplets with all relation types. However, each sentence may contain very few relation types. In this paper, we first model the final triplet extraction result as a three-order tensor of word-to-word pairs enriched with each relation type. And in order to obtain the sentence contained relations, we introduce an independent but joint training relation classification module. The tensor decomposition strategy is finally utilized to decompose the triplet tensor with predicted relational components which omits the calculations for unpredicted relation types. According to effective decomposition methods, we propose the {\bf T}ensor {\bf D}ecomposition based {\bf R}elation {\bf E}xtraction (TDRE) approach which is able to extract overlapping triplets and avoid detecting unnecessary entity pairs. Experiments on benchmark datasets NYT, CoNLL04 and ADE datasets demonstrate that the proposed method outperforms existing strong baselines.

\end{abstract}


\end{frontmatter}

\section{Introduction}

Relation Extraction (RE) is aiming to extract entity pairs with relation types from unstructured text, which facilitates many other Natural Language Processing (NLP) tasks, including knowledge base construction and question answering. For example, there is a given sentence ``Bill Gates co-founded Microsoft with his friend Paul Allen", and the task is to extract triplet (Bill Gates, Founder, Microsoft). Traditional pipeline models consist of two separate subtasks name entity recognition and relation classification, which first extract entities and then classify entity pairs with relation types for decoding these into triple formats. However, pipeline models suffer from error propagation and do not fully use the relevance of these two subtasks. To tackle these problems, most recent studies pay attention to joint models which integrate entity recognition and relation classification into a single model by jointly training like training paradigm in multi-tasks projects. Owing to reasonable information sharing, joint models have achieved better performance than the pipeline models. 

Most existing joint models rely on fine-grained labeling scheme~\cite{suncong-zheng:novel-tagging} or train by  parameter sharing~\cite{sun-etal-2019-joint,fu-etal-2019-graphrel} in the embedding layers. However, fine-grained labeling scheme is unable to identify overlapping relations.  As an improvement, the authors model triplet extraction as multi-head selective problems in \cite{DBLP:journals/corr/abs-1804-07847}, which approximately calculates the joint probability $p(e_s, r, e_t)$ where $(e_s, r, e_t)$ is a triplet in the given sentence, $e_s$ denotes the source entity, $e_t$ denotes the target entity and $r$ denotes the relation between the source entity and the target entity. Having these notations in mind, we could model these predicted approximate joint probabilities as an $N \times N \times K$ tensor, where $N$ is the number of words in the sentence $s$ under consideration and $K$ is the number of predefined relation types. Each element in this tensor denotes the probability of the existence of a relation among word pairs. However, the great influence of relation classification would be ignored if we directly model the joint probabilities. And it could result in predicting unnecessary triplets when the predicted relations are wrong. As is well known, the triplet is correct only when all elements in the triplet are predicted correctly, which relies on the corrected predicted relations. Wrong predicted relation types could decrease the accuracy of the extracted triplets and cost a lot of unnecessary calculations. 

In this paper, we attempt to avoid predicting redundant entity pairs while extracting overlapping triplets. It also means to prune the extracted tensor that could decode overlapping relations. Based on the tensor notation mentioned above, we consider the final extracted triplets (joint probability) as a three-dimensional tensor. By introducing a tensor based decomposition into directional component (DEDICOM) algorithm \cite{article:dedicom}, we decompose the extracted triplets tensor into relational component so that it can effectively capture the interactions among relation types. From the perspective of probability, decomposing  $p(e_s, r, e_t|s) = p(e_s, e_r|s, r)p(r|s)$ is an effective way to simplify the original joint probabilities. To extract multiple kinds of relation types $p(r|s)$ that the sentence contains, we regard relation classification model as an independent module but also a part of joint training process. And then we define a new operation between the diagonal tensor appeared in the DEDICOM algorithm and the result of relation classification to restrict the detection of entity pairs with no prediction relation types. It is able to avoid  extracting redundant triplets. Originally, the tensor based DEDICOM strategy is able to capture the asymmetric interactions among word pairs in the sentence. This idea is suitable for our task because the extracted triple tuple $(e_s, r, e_t)$ is directional where $e_s$ is the source entity and $e_t$ is the target entity.

The main work of this paper includes:
\begin{enumerate}
    \item We propose a new joint learning framework with tensor decomposition strategy for relation extraction (TDRE). Additionally, we employ the DEDICOM strategy for capturing the interactions among relation types.
    \item We introduce a new joint training relation classification module and apply the results into tensor decomposition process,  which is able to avoid extracting redundant triplets.
    \item We conduct extensive experiments on three benchmark datasets NYT10, CoNLL04 and ADE datasets, which show that our model is able to achieve better performance than most of the existing strong baseline models.
\end{enumerate}

\section{Related Work}

Many previous relation extraction approaches focus on manual feature extraction which heavily rely on NLP tools and sometimes are only applicable to specific fields. Recent years, neural networks and deep learning have attracted more and more attention. Deep neural network models have made significant progress in relation extraction without complicated handcrafted features. The task of relation extraction can be divided into two categories: pipeline models~\cite{Fundel-Clemen:relex,extraction:medical-case} and joint models~\cite{miwa-bansal-2016-end,suncong-zheng:novel-tagging,DBLP:journals/corr/abs-1804-07847,sun-etal-2019-joint}. Pipeline models take the task as two separated models including name entity recognition model and relation classification model. However, separated models cannot dig the potential relevance between two subtasks and suffer from error propagation. Joint models take these subtasks into a single model and train jointly by shared parameters. 

Zheng et al. \cite{suncong-zheng:novel-tagging} introduced a novel tagging scheme where each word was tagged with a unique tag including an entity type and a relation type. Hence, the set of word tags is the Cartesian product of the relation types and the entity types. However, it is unable to deal with overlapping relations which means that one word may be mapped with many other words with different relations in the text. Zeng et al. \cite{zeng-etal-2018-extracting} proposed a sequence-to-sequence copy mechanism to decode overlapping triplets. Bekoulis et al. \cite{DBLP:journals/corr/abs-1804-07847} considered entities-relation extraction tasks as multi-head selection problems which could effectively solve overlapping relations problem as well. With the development of graph convolution neural network, it is developed to capture not only the ordered features on the timeline but also the features between different nodes in space with graph representation. Sun et al. \cite{sun-etal-2019-joint} made fully uses of the relevance between entity and relation types by building entity-relation graph to catch the combination of entity pairs and valid relation. Guo et al. \cite{guo-etal-2019-attention} also used graph convolution networks to capture structured information with dependency trees and to select relevant sub-structures with a soft-pruning approach. Nevertheless, this kind of proposed methods predict each relation for every word-to-word pairs which still suffer from redundant triplets. Takanobu et al. \cite{HRL} firstly detected relation types, and then identified related entity pairs with hierarchical reinforcement learning strategy which avoided predicting redundant relation types to some extent.

Inspired by the aforementioned proposed methods, we introduce an independent but joint training relation classification module. And based on the DEDICOM strategy, we develop a joint learning framework for relation extraction with tensor decomposition algorithms. The proposed tensor decomposition based relation extraction model is effective to avoid predicting redundant triplets.

\section{Model}

\begin{figure*}[h]
\centering
\includegraphics[height=9cm, width=17cm]{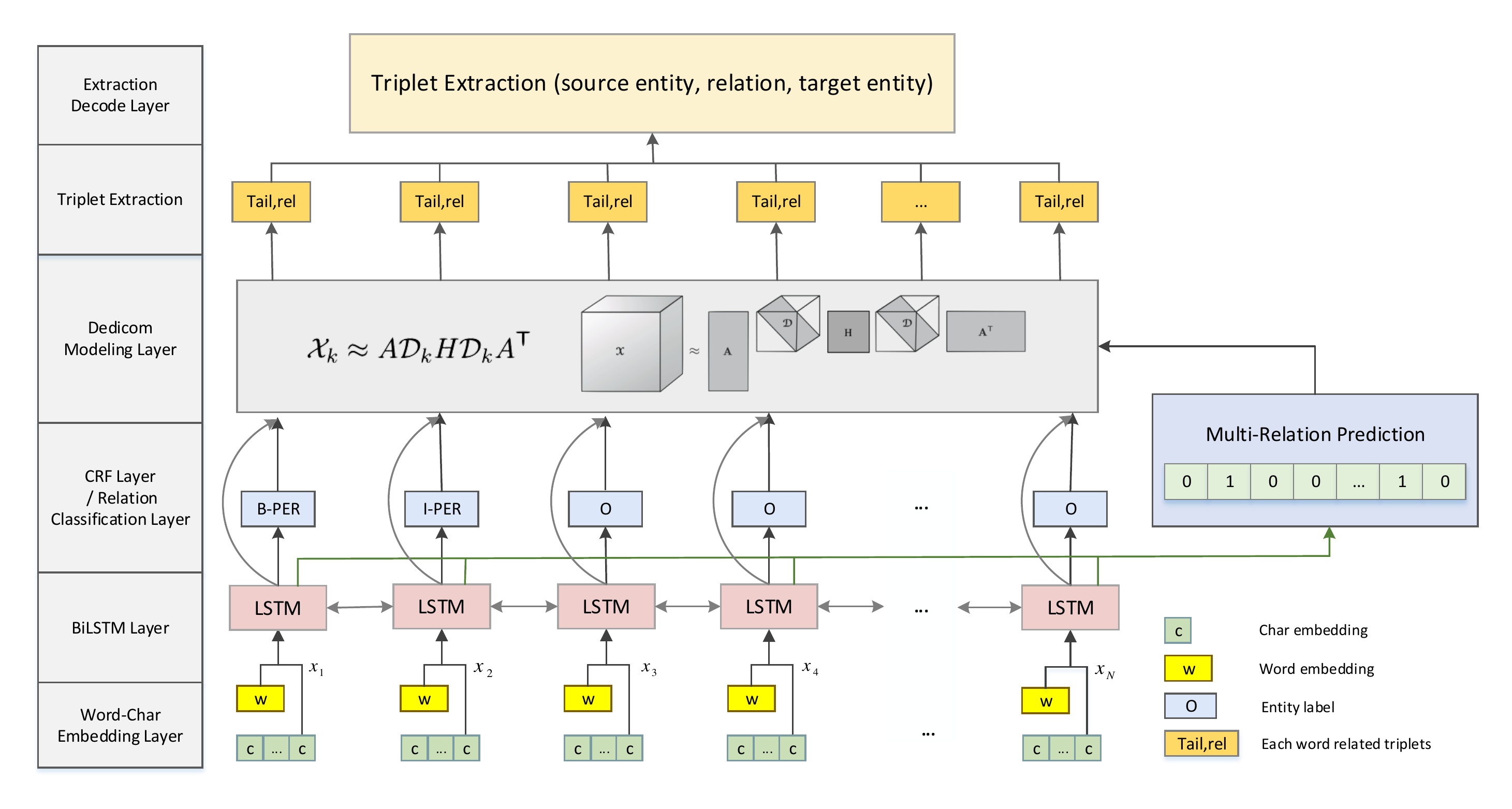}
\caption{The overall architecture of the proposed tensor decomposition based model for relation extraction. }
\label{figure-overall}
\end{figure*}

In this section, we present our tensor decomposition based relation extraction model (TDRE) which is illustrated in Figure \ref{figure-overall}. 

The relation extraction problem is denoted as follows. Given the relation types set $\mathcal{R} = R \cup \{\text{``N"}\}$, where $R$ represents the dataset of predefined relation types and ``N" for none-relation types. If the word is not matched with any other words, it will be assigned the relation type of ``N". Given a sentence $S$ consisting of $n$ words $w_1w_2...w_n$, what we need is to structure the text into multiple triple tuples $(e_s, r, e_t)$, where $e_s$ denotes the source entity, $e_t$ denotes the target entity and ${r}$ denotes the relation between the source entity and target entity.

\subsection{Embedding Layer and BiLSTM layer}
In order to map each word into a word vector, we use pre-trained word embeddings with the skip-gram word2vec model~\cite{Mikolov:2013:DRW:2999792.2999959}. And we combine character embeddings with convolutional neural networks (CNN) since they are able to capture morphological features such as prefixes and suffixes. Bidirectional Long Short-Term Memory Networks (BiLSTM) is utilized to encode bidirectional contextual information~\cite{Graves2013Speech}, 
\begin{align}
h_i = \text{BiLSTM}(x_i, h_{i-1}),
\end{align}
where $h_i$ represents the hidden states with BiLSTM networks concatenating the forward and backward hidden states at position $i$ in the given sequence, and $x_i$ represents the $i$th word embedding representation.  Here, we formally express $x_i$ as $[\text{word\_emb}(w_i); \text{CNN}(\text{char\_emb}(w_i))]$ by pre-trained word vector embeddings and CNN based character embeddings. For this layer, we will get a final representation matrix $A$ for the given sentence:
\begin{align} \label{lstm_out}
A = [h_1, h_2, ..., h_n] \in \mathbb{R}^{n \times d_h},
\end{align}
where $n$ is the sequence length and $d_h$ is the final concatenated bidirectional hidden dimension of BiLSTM networks.

\subsection{Name Entity Recognition Model}

To identify the entities in a given sentence, we regard entity recognition task as a sequence labeling problem as what is done in \cite{DBLP:journals/corr/HuangXY15}. A Conditional Random Fields (CRF) layer is employed here. The designed state transition matrix makes full use of neighbor tag information, which helps both entity type identification and boundary recognition. With the sentence representation $A$ after embedding layer and BiLSTM layer,  the given entity tag set $ \{y_1, y_2, ..., y_m\}$ and the defined feature score functions ${F_1, F_2, ..., F_J}$, the probabilistic model is formulated as:
\begin{equation}
 \label{crf_eq}
p(y_k|h_i) = \frac{ \exp( \sum_{j=1}^J \lambda_j F_j(y_k, h_i))}{\sum_{y^{'}} \exp( \sum_{j=1}^J \lambda_j F_j(y^{'}, h_i))},
\end{equation}
where $i=1, 2, ..., n; k = 1, 2, ..., m$ and $\lambda_j$ shows each feature function may have different weight to decide the final probabilistic score. Each feature function $F_j$ may include several sub-feature functions $f$ which can be formalized as $F_j = \sum_if_j(y_{i-1}, y_i, A, i)$ where $i$ represents the position of current word in the given sentence, $y_i$ denotes the entity tag of current word and $y_{i-1}$ denotes the entity tag of the previous word. With these definitions, the loss function is formally given as:
\begin{align}
 L_{ner} & = - \log p(y|A) \\
    & = - \sum_{j=1}^{J} \lambda_j F_j(y, A) + \log Z(\lambda, A),
\end{align}
where $y$ is the ground truth word label and $Z(\lambda, A) = \sum_{y^{'}} \exp(\sum_{j=1}^J \lambda_j F_j(y^{'}, A))$. Our optimization goal is to minimize the loss function and finally decode entity tags with Viterbi algorithm.

\subsection{Relation Classification Model}

As we all know, simple classification problem is much easier than simultaneous classification for both relations and its related entity pairs. In this module, we consider the relation classification problem as an independent multi-label classification problem, because each sentence may contain more than one relation type~\cite{RCNN,Liu:2016:RNN:3060832.3061023,yang-etal-2016-hierarchical,GCN_TC}. In order to share the word representation, we use the outputs of BiLSTM networks as the inputs of our classification model which could help find the interactions between entity recognition and relation classification. It is worth noting that we are categorizing for the entire sentence here instead of categorizing every word in the sentence.  

A linear network is utilized in our relation classification model and it can be denoted as $f(A)= WA + b$, where $A$ is the sentence representation, $W$ and $b$ are the training parameters.  As for the multi-label problems, we focus on the final expression of the classification results $P = (P_1, P_2, ... P_K)$ and each element is calculated as follows: 
\begin{align}
\label{cal_rel_cls}
P_k = \begin{cases}1, & \text{ if $\sigma(f(A)) \geq \gamma_1$} ;\\
0, & \text{else if $\sigma(f(A)) < \gamma_1$}.
\end{cases}
\end{align}
where $\sigma$ is the sigmoid activation function and $\gamma_1$ is the threshold for predicting the sentence contained relation types.

If the real relation types are denoted by $y$, then the goal of this classification module is to minimize the loss function:
\begin{align}
L_{rel} = -\sum_{k=1}^{K}y_k \log p_k, 
\end{align}
where $y_k$ is the ground truth relation type and $p_k$ is the probability for predicting the current sentence into the $k$th relation type. And in the training process, sigmoid cross-entropy loss function is used since there exists a multi-label problem.

\subsection{Triplet Extraction Model}

According to the above defined sentence with $n$ words, the target of relation extraction is to extract the existing triple tuples with the predefined relation type set $\mathcal{R}$ from the unstructured text. And the extracted triplet $(e_s, r, e_t)$ corresponds to specified entity spans $e_s, e_t$ and the relation $r$ between $e_s$ and $e_t$. For a specific relation type $k$,  we focus on mapping each word with the others in the sentence with an indication function. The word-to-word pair mapping can be modeled as a matrix $X \in {\mathbb{R}^{n \times n}}$, where the element $x_{ij}$ in the matrix $X$ indicates whether the $i$th word and the $j$th word can form a triplet related word pair. And $i$ and $j$ represent the last word of source entity and target entity respectively. Hence, the indication function is expressed as $x_{ij} = 1$ when the $i$th word can map with $j$th word in the specific relation type $k$. Extending matrix $X$ with relation type component, we have a tensor extraction representation $\mathcal X$. It can be defined as following:
\begin{align} \label{tensor_define}
\mathcal X_{ijk} = \begin{cases} 1, & \text {if there is a triplet $(i, k, j)$; }\\
0, & \text{otherwise}.
\end{cases}
\end{align}
Note that in this mathematical formula, $i$ and $j$ are the tail words of two entity spans. 

In order to extract useful information from the tensor, we employ a tensor decomposition algorithm named decomposition into directional component (DEDICOM) strategy introduced by Harshman et al. \cite{article:dedicom}. As for a sequence with $n$ words, the defined mapping matrix $X \in {\mathbb{R}^{n \times n}}$ needs to describe the asymmetric relationships between each word and the other words. This idea is suitable for that each extracted triplet is directional since source entity and target entity are ordered. Triplet extraction process can be modeled as a three-order DEDICOM model, where the target constructed tensor is $\mathcal{X} \in {\mathbb{R}^{n \times n \times K}}$, where $K$ denotes the number of relation types. The decomposition is:
\begin{align}
 \label{dedicom-eq}
\mathcal{X}_k \approx A\mathcal{D}_kH\mathcal{D}_kA^{\mathsf{T}}  & \text{    for $k=1, ..., K$},
\end{align}
where $A \in \mathbb{R}^{n \times  d_h}$ is the sequence representation after BiLSTM networks, $\mathcal{D} \in \mathbb{R}^{d_h \times d_h \times K}$ and $H \in \mathbb{R}^{d_h \times d_h} $ are parameters. The matrices $\mathcal{D}_k \in \mathbb{R}^{d_h \times  d_h}$ are diagonal, and the diagonal entry $(\mathcal{D}_k)_{ii}$ indicates the participation of $i$th latent component at relation $k$ and $i = 1, 2, ..., d_h$.

From the above mentioned module, it is necessary to predict all relation types for each word pair while the sentence may contain only few relation types. To address this problem and to further utilize the predicted relation types with relation classification model, we need to make specific settings for the factor of $\mathcal{D}$  in the structure of the tensor decomposition. Actually, it is not necessary to calculate the mapping matrix $\mathcal{X}_r$ when there is no prediction for the $r$th relation type. We set $\mathcal{D}_r^{'}$ as a zero matrix, thus $\mathcal{X}_r$ turns into a zero matrix after calculation, which means no triplet in $r$th relation component. Thus, it is theoretically observable that the decomposition strategy avoids predicting unnecessary triplets and simultaneously reduces the predicting time. 
Therefore, a custom operation $\circ$ can be represented as follows:
\begin{align} \label{cal_d_k}
\begin{small}
\mathcal{D}_{k}^{'} = \mathcal{D}\circ P_k = \begin{cases} \text{diagonal matrix, } & \text {if $P_k=1$; }\\
\text{zero matrix, } & {\text{else if } P_k = 0}.
\end{cases}
\end{small}
\end{align}
And then, the final decomposition algorithm can be formalized as:
\begin{equation} 
\label{dedicom_p}
\begin{small}
\mathcal{X} \approx A (\mathcal{D}\circ P) H(\mathcal{D} \circ P) A^{\mathsf{T}} ,
\end{small}
\end{equation}
where $P \in \mathbb R^K$ denotes the prediction result of relation classification module. 

From the perspective of probability distribution, we do not directly model the joint probability like what is done in \cite{DBLP:journals/corr/abs-1804-07847} and \cite{fu-etal-2019-graphrel} but decompose joint probability into relation component as conditional probability:
\begin{align}
p(x_i, r_k, x_j|S) = p(x_i, x_j|r_k, S)p(r_k|S),
\end{align}
$i, j=1, 2, ..., n; k=1, 2,..., K$, where $p$ represents probability distribution, $x_i, x_j$ respectively denotes the $i$-th and $j$-th word representation in the sentence $S$ and $r_k$ denotes the $k$-th relation type. Owing to that each word may be mapped with more than one word and relation types, we employ sigmoid function here to amplify the difference. And the final triplet decode procession can be denoted as:
\begin{align}
\label{triplet_decode}
P(x_i, r_k, x_j|S) = 
\begin{cases}
1, &  \sigma(\mathcal X_{ijk}^{'}) \geq \gamma_2; \\
0, &  \sigma(\mathcal X_{ijk}^{'}) < \gamma_2 .
\end{cases}
\end{align}
where $\mathcal X_{ijk}^{'} $ is the calculated tensor and $\gamma_2$ represents the threshold for judging whether a triplet is true.

The goal of this decomposed relation extraction module is to minimize the cross-entropy loss $L_{extract}$ during training:
\begin{align}
L_{extract} = -\sum_{i, j, k}  \mathcal X_{ijk}\log \mathcal X_{ijk}^{'} ,
\end{align}
where $\mathcal X_{ijk}$ is the ground truth label and $\mathcal X_{ijk}^{'} $ is the predicted triplet tensor.

\begin{algorithm}[tb]
\caption{Tensor Decomposition for Relation Extraction (TDRE)}
\label{alg:algorithm}
\textbf{Input}: Sentence $S$, predefined relation type set $R$\\
\textbf{Output}: Triplets $\{(e_s, r, e_t)_{j}\}_{j=1}^{m}$, $e_s, e_t$ are the source entity and the target entity,  $r$ denotes the relation type, and $m$ denotes the number of triplets.
\begin{algorithmic}[1]
\STATE Define length of the given sentence: $n$, number of relations: $K$.
\REPEAT
\STATE Calculate word representation $A$ by Eq. \eqref{lstm_out}; \\
\STATE Identify entity types $Y_{ner}$ by Eq. \eqref{crf_eq} ;

\IF {entity label embedding}
\STATE $A \leftarrow A + \text{label\_embedding}(Y_{ner})$
\ENDIF
\STATE Initialize decomposition factor $\mathcal D$ and $H$; \\
\IF {relation classification model}
\STATE Classify relation types $P$ by Eq. \eqref{cal_rel_cls};
\STATE Calculate diagonal relation matrix $\mathcal D$ by Eq. \eqref{cal_d_k};
\STATE Calculate triplet tensor $\mathcal X_{ijr}$ by Eq. \eqref{dedicom_p};
\ELSE 
\STATE Calculate triplet tensor $\mathcal X_{ijr}$ by Eq. \eqref{dedicom-eq};
\ENDIF
\STATE Decode triplets with Eq. \eqref{triplet_decode} and Eq. \eqref{tensor_define}.

\UNTIL{ TDRE converges} 
\end{algorithmic}
\end{algorithm} 

\subsection{Loss Function}

In order to jointly train the proposed model, we combine all of the three objective loss functions and optimize parameters together. The final loss function is computed as follows:
\begin{align}
L = L_{ner} + L_{rel} + L_{extract},
\end{align}
where $L_{ner}$ is the loss function in name entity recognition module, $L_{rel}$ is the loss function in relation classification module and $L_{extract}$ is the loss function of final triplet extraction module which is calculated by DEDICOM algorithm. In order to update parameters, we optimize our model with Adam optimization approach and train the proposed joint model like multi-tasks learning paradigm.

Finally, our TDRE algorithm is summarized in Algorithm \ref{alg:algorithm}.

\section{Experiments}

\subsection{Datasets}

We conduct experiments on three benchmark datasets for relation extraction: (i) NYT10, the originally New York Times corpus~\cite{Riedel} which is developed by distant supervised and published by Takanobu et al. \cite{HRL} who filtered dataset by removing the relations in training set but not in testing set and sentences containing no relations. (ii) CoNLL04 dataset~\cite{roth-yih-2004-linear}  and it is split by Gupta et al. \cite{gupta-etal-2016-table} and Adel et al. \cite{adel-schutze-2017-global}. The official given entity type set is \{Location, Organization, Person, Other\} where we omit the beginning, inside and the outside tags of entity types here. And the official defined relation type set is \{{Kill, Live in, Located in, OrgBased in, Work for\}. 
(iii) Adverse Drug Events dataset (ADE)~\cite{Gurulingappa2012DevelopmentOA}, where we use 80\% for training and 20\% as test set. Entity type set \{Beginning, Inside, Outside\} is applied here since there is no official entity types. And the relation type set is \{Adverse-Effect, Drug-Disease Treatment\}. 

The details of the public datasets are reported in Table \ref{tab:statistics}. The approximate statistical number of triplets on ADE dataset is shown since we conduct cross-validation on ADE dataset.

\begin{table}[h]
\caption{Statistics of the datasets}
\label{tab:statistics}
\centering
\begin{tabular}{lccc}
\hline
Datasets  &NYT10&CoNLL04&ADE\\
\hline
Relation types     &29&5&2   \\
Entity types    &7&4&3     \\
Training set   &70339&910&3416    \\
Training triplets    &87739&1273&5650   \\
Test set       &4006&288&854   \\
Test triplets    &5859&422&1450  \\
\hline
\end{tabular}
\end{table}

\subsection{Details of Implementation}

Similar to previous works, we obtain the 50-dimensional word embeddings which is used by \cite{adel-schutze-2017-global} trained on Wikipedia. Character embeddings are initialized randomly and the kernel size is set to 3 for convolution neural network while parameters update with training process. And then we concatenate the word embeddings and character embeddings as the final input embeddings. For the representation of the learning layers, we use three-layer BiLSTM networks with 64 hidden units. Comparing to most existing models, we randomly initialize entity label embeddings for all conducted datasets with embedding size 128, and update parameters by optimization. Training is performed by using the Adam optimizer with learning rate $10^{-3}$, and dropout technique is used in input embeddings and BiLSTM hidden layers, where the dropout rate is set to 0.1. Early stopping is also used on the validation set for avoiding overfitting.

\begin{table*}[h]
\caption{Main results on NYT10 and NYT10-sub datasets where NYT10-sub is selected from NYT10 test set. And all baseline results come from the original papers of HRL~\cite{HRL}.}
\label{tab:dataset_nyt}
\centering
\begin{tabular}{lcccccc}
    \toprule
    \multirow{2}{*} {-Methods} &  \multicolumn{3}{c} {NYT10} &  \multicolumn{3}{c} {NYT10-sub} \cr
     \cmidrule(lr){2-4} \cmidrule(lr){5-7} 
       &Pre&Rec&F1&Pre&Rec&F1\cr
    \midrule
    Tagging \cite{suncong-zheng:novel-tagging} &{59.3}&{38.1} & {46.4}&{ 25.6} & { 23.7 } & { 24.6 } \cr
    CopyR \cite{zeng-etal-2018-extracting}&56.9 & 45.2& 50.4& 39.2 & 26.3 & 31.5  \cr
    HRL \cite{HRL}&71.4 & 58.6 & 64.4&  81.5 & 47.5  &  60.0  \cr
     MrMep \cite{Mrmep} & 71.7 & {63.5} & 67.3 & {\bf 83.2} & 55.0 & 66.2  \cr
     TDRE (ours) & {\bf 81.3} & {\bf 68.3 } & {\bf 74.3 } & { 82.2} & {\bf 70.2} & {\bf 75.7} \cr
    \bottomrule
    \end{tabular}
\end{table*}

\subsection{Results}
\paragraph{Evaluation Metrics} As for entity recognition, the entity is correctly identified only when all the characters of the original entity are recognized and recognized as the correct type. As for triplet extraction, a triplet is considered correct if and only if the source entity, target entity and related relation types are  both correct. We adopt precision, recall rate and micro-F1 to evaluate the performance.

\paragraph{Baselines} For comparison, we choose the following models as baselines: 
\begin{enumerate}
\item Tagging \cite{suncong-zheng:novel-tagging}. It takes the joint extraction as a sequential labeling problem with a special tagging schema where the label of each word includes the entity type and entity order (source or target entity) and the relation type.
\item CopyR \cite{zeng-etal-2018-extracting}. It is a sequence-to-sequence learning framework with copy mechanism which could handle the relational triplet overlapped problem.
\item HRL \cite{HRL}. It applies a hierarchical reinforcement learning by detecting relations and then extracting participating entities for the relation which regards the related entities as the argument of a relation.
\item MrMep \cite{Mrmep}. It is also a joint learning model which first classifies relations and then uses triplet attention to reinforce the connections between relation and entity pairs.
\item Multi-head selection model~\cite{DBLP:journals/corr/abs-1804-07847}. It takes the relation extraction task as a multi-head selection problem, which could identify multiple relation types for each entity pair. 
\item Multi-head with adversarial training regularization method~\cite{bekoulis-etal-2018-adversarial}. It has also achieved better performance in overlapping relations. 
\item Li et tal. \cite{Li2017ANJ} introduces a neural joint model to simultaneously extract both biomedical entities and their relations by using hand-crafted features or features derived from NLP tools. 
\item MultiQA~\cite{li-etal-2019-entity}. It takes the task as a multi-turn question answering problem which identify answer spans (entities) from the defined relation related questions. 
\end{enumerate}

\paragraph{Experiment Results} The detailed comparisons between baseline models and our proposed model are summarized in Table \ref{tab:dataset_nyt} and Table \ref{tab:result-conll04-ade} with precision, recall, micro-average F1. The best performances are marked in bold. 

\begin{table*}[ht]
\caption{Main results of  our method comparing with previous baseline models on benchmark dataset. Overall F1 is to average these two subtasks. Results of the compared baseline models directly come from the original papers.}
\label{tab:result-conll04-ade}
\centering
\begin{tabular}{llccccccc}
    \toprule
    \multirow{2}{*} {Dataset}&\multirow{2}{*}{Model}&
    \multicolumn{3}{c}{Entity Recognition}&\multicolumn{3}{c}{Triplet Extraction} &\multirow{2}{*} {Overall F1}\cr
    \cmidrule(lr){3-5} \cmidrule(lr){6-8}
    &&Pre&Rec&F1&Pre&Rec&F1&\cr
     \midrule
     \multirow{4}{*}{CoNLL04}
    & Multi-Head & 83.75 & 84.06 & 83.9 & 63.75 & 60.43 & 62.04 & 72.97 \cr
    & MultiHead-AT & - & - & 83.61 & - & - & 61.95 & 72.78 \cr
    & MultiQA & 89.0 & 86.6 & 87.8 & 69.2 & 68.2 & 68.9 & 78.35\cr
    &{TDRE (ours)} &{\bf 91.85}&{\bf 91.34}&{\bf 91.59}&{\bf 80.90}&{\bf 76.67}&{\bf 78.73}& {\bf 85.16} \cr
    \midrule
     \multirow{4}{*}{ADE}
    & Li \cite{Li2017ANJ} & 82.70 & 86.70 & 84.60 & 67.50 & 75.80 & 71.40 & 78.00 \cr
    & Multi-Head& 84.72 & { 88.16} &86.40 & 72.10 & {\bf 77.24} & 74.58 & 80.49 \cr
    & MultiHead-AT& - & - & 86.73 & - & - & 75.52 & 81.13 \cr
    & {TDRE (ours)} & {\bf 88.82} & {\bf 88.28} & {\bf 88.55} &{\bf 81.63 }& { 76.20} & {\bf 78.82} & {\bf 83.69} \cr
    \bottomrule
    \end{tabular}
\end{table*}

As for NYT10 dataset, the purposed TDRE model significantly outperforms the others. Through the sub test set NYT10-sub published by HRL \cite{HRL}, it is well known that it contains most of the overlapping relations while the triplets shared both source entity and target entity. Our model shows the best performance particularly in NYT10-sub which means TDRE is more effective for overlapping problems. And in Table \ref{tab:result-conll04-ade},  it can be seen that our proposed approach performs better comparing with the baseline methods whatever in entity recognition or relation identification on CoNLL04 and ADE dataset. 

Especially focusing on the precision of triplet extraction, experimental results show great improvements. Theoretically, our model reduces extra computation for the unpredicted relation types while most of the other models compute every relation type for an entity pair. The agreement between theoretical analysis and experimental results show that the model can indeed reduce predicting the redundant triplets. All the experimental results indicate that the proposed approach of tensor decomposition into relation component to model relation extraction process is more effective than the baseline models, even without using BERT. 

\paragraph{Ablation Study} In order to illustrate the effectiveness of various parts of our model, we conduct ablation experiments on the CoNLL04 dataset. Concretely, we separately remove character embeddings (Char Emb.), entity label embeddings (Label Emb.) and relation classification module (CLS). According to the results presented in Table \ref{tab:ablation-study}, it can be observed that character embeddings, which could catch additional information such as suffixes, prefixes and other morphological features, are helpful in the triplet extraction. In terms of entity label embeddings, it is possible to incorporate entity type information and boundary information which is beneficial to capture the integrity of entities. As for removing classification module,  the proposed approach degenerates in directly modeling joint probability and the performances severely degrade whether in entity recognition or in triplet extraction. This indicates the proposed tensor decomposition into relation component approach has great contributions in triplet extraction.
 \begin{table}
 \caption{Ablation study on CoNLL04 dataset.}
\label{tab:ablation-study}
\centering
\begin{tabular}{lccccc}
    \toprule
    \multirow{2}{*} {Methods} & {Entity} &  \multicolumn{3}{c} {Triplet Extraction} \cr
     \cmidrule(lr){2-2} \cmidrule(lr){3-5} 
       &F1&Pre&Rec&F1\cr
    \midrule
    TDRE &{\bf91.59}&{\bf 80.90 } & {\bf 76.67} & {\bf 78.73} \cr
    - Char Emb. &85.65& 75.91 & 64.52 & 69.75 \cr
    - Label Emb. &89.51& 75.74 & 66.90 & 71.05 \cr
     - CLS &88.80& 74.53 & 66.90 & 70.51 \cr
    \bottomrule
    \end{tabular}
\end{table}

 \paragraph{Analysis of Relation Classification Model} Experiments have been conducted on CoNLL04 dataset with recurrent neural networks (RNN), recurrent convolution neural networks (RCNN) and bidirectional long short-term memory networks (BiLSTM) models for relation classification. The results in Table \ref{tab:cls-study} show that our model are still better than the previous compared models \cite{DBLP:journals/corr/abs-1804-07847,li-etal-2019-entity} even without the classification model. We guess it is because the raw introduced DEDICOM strategy decomposes with diagonal tensor, and it is to interpret connections between relation types. The strategy is originally suitable for capturing asymmetry which is fit to our ordered triplets. Focusing on the other results with RNN and RCNN, performances are better than multi-head model but not better than MultiQA~\cite{li-etal-2019-entity}. There are two possible reasons: (1) classification models pay much attention on identifying relations with sacrificing the accuracy of entity recognition; (2) parameters are not shared with name entity recognition in these classification models, which ignores the inherent connections between relation identification and entity recognition. After parameters sharing in BiLSTM networks, the model achieves substantial improvements in entity recognition and triplet extraction module. Better performance in relation identification or in entity recognition can yield better triplet extraction performance.
  
 \begin{table}[h]
 \caption{Various classify models on CoNLL04 dataset.}
\label{tab:cls-study}
\centering
\begin{tabular}{lccccc}
    \toprule
    \multirow{2}{*} {Model} & {Rel} & {Entity} & \multicolumn{3}{c} {Triplet Extraction} \cr
    \cmidrule(lr){2-2} \cmidrule(lr){3-3} \cmidrule(lr){4-6}
     &F1&F1&Pre&Rec&F1\cr
    \midrule
    - CLS& - & 88.8 & 74.5 & 66.9 & 70.5\cr
    RNN & 97.6 & 85.0 & 70.5 & 60.2 & 64.9 \cr
    RCNN & {\bf 99.7} & 87.7 & 75.2 & 66.4 & 70.5 \cr
    \midrule
    BiLSTM & 94.1 &{\bf 91.6}&{\bf 80.9 } &{\bf 76.7 } &{\bf 78.7 }\cr
    \bottomrule
    \end{tabular}
\end{table}

\section{Conclusion} 

In this paper, we introduce a joint neural model with tensor decomposition based strategy for multi-labeled relation extraction (TDRE) . The introduced tensor is to model the connections of word pairs in each relation component, which could tackle overlapping relations. Consider relation classification as an independent but joint training module to get sentence contained relation types. By implanting predicted relations to the tensor decomposition formula as relational component factors, it can avoid identifying unnecessary entity pairs among no related relation types. Experiments on public datasets demonstrate that the proposed model achieves significant improvements over previous state-of-the-art baselines. In the future, we plan to explore other tensor decomposition approaches and extend our model in other tensor based modeling tasks.

\section*{Acknowledgments}
This work is supported by the Science and Technology Plan Project of Sichuan Province (Key R \& D Project, 2020YFS0465) and Mathematics Teaching Research and Development Center for Colleges (CMC20190501).

\bibliographystyle{elsarticle-num}
\bibliography{TDRE}

\end{document}